\documentclass[runningheads]{llncs}
\usepackage{graphicx}
\usepackage{booktabs}
\usepackage[colorinlistoftodos]{todonotes}
\usepackage[inline]{enumitem}
\usepackage[hidelinks]{hyperref}
\usepackage{physics}

\newcommand{\citep}[1]{\cite{#1}}
\newcommand{\setcard}[1]{|{#1}|}

\usepackage{chngcntr}
\counterwithout{figure}{chapter}
\counterwithout{table}{chapter}

\begin{document}
\title{Does Wikidata Support Analogical Reasoning?}

\author{Filip Ilievski \and Jay Pujara \and Kartik Shenoy}
\authorrunning{Ilievski et al.}
%
\institute{Information Sciences Institute, University of Southern California\\
\email{\{ilievski, jpujara, kshenoy\}@isi.edu}}

%
\maketitle              
\setcounter{footnote}{0}
\setcounter{table}{0}

\begin{abstract}
Analogical reasoning methods have been built over various resources, including commonsense knowledge bases, lexical resources, language models, or their combination.
While the wide coverage of knowledge about entities and events make Wikidata a promising resource for analogical reasoning across situations and domains, Wikidata has not been employed for this task yet. In this paper, we investigate whether the knowledge in Wikidata supports analogical reasoning. Specifically, we study whether relational knowledge is modeled consistently in Wikidata, observing that relevant relational information is typically missing or modeled in an inconsistent way. Our further experiments show that Wikidata can be used to create data for analogy classification, but this requires much manual effort. To facilitate future work that can support analogies, we discuss key desiderata, and devise a set of metrics to guide an automatic method for extracting analogies from Wikidata.
\end{abstract}

\section{Introduction}
\label{sec:intro}

Cognitive science research has provided rich evidence that humans use analogical reasoning to understand, explain, or imagine novel situations within or across domains~\cite{holyoak1995mental}. Analogical thinking can connect the Great Depression and the financial crisis based on \textit{causal} knowledge~\cite{nagarajah2022understanding}, or compare the Sun and the Earth to the Earth and the Moon based on the \textit{revolves} relation~\cite{gentner1997analogy}. Corresponding cognitive systems have been build to realize and test this skill algorithmically, such as the Structured Mapping Engine~\cite{Forbus2017ExtendingST} and the Companion architecture~\cite{Forbus2017AnalogyAQ}. Natural Language Processing research on analogy has been popularized through the proportional analogy task, illustrated through the famous example of \textit{man:woman-king:queen} by the word2vec system~\cite{mikolov2013linguistic}. Recognizing the gap between the large-scale word analogy systems and the expressive cognitive systems, recent AI research has focused on integrating neural (language) models with cognitive systems to solve tasks like sketch object recognition~\cite{chen2019human}, product innovation~\cite{hope2017accelerating}, narrative understanding~\cite{nagarajah2022understanding}, and moral decision making~\cite{dehghani2008integrated}.


As it can be expected, these analogical reasoning efforts are often centered around a knowledge base that enables models to understand implicit relations, such as \textit{causes} or \textit{revolves}. Curiously, despite the large quality and richness of Wikidata, and its increasing adoption for many knowledge-intensive tasks~\cite{moller2021survey,klein2021generating,oguz2020unik}, prior work on analogical reasoning has not considered leveraging its ontology or its factual knowledge to reason by analogy. Instead, existing systems have typically leveraged publicly available parts of Cyc, semantic lexical resources, language models~\cite{ushio-etal-2021-bert}, or their combination~\cite{Forbus2017AnalogyAQ}. 

Considering the coverage of millions of ontological concepts and instances, intuitively, Wikidata could serve as a valuable resource for analogical reasoning. In this paper, we perform an initial study on whether the Wikidata knowledge supports analogical reasoning. Specifically, we focus on three key questions:
\begin{enumerate}
    \item \textit{Does Wikidata express relational information consistently?} Analogical reasoning revolves around relational similarity, therefore, consistency in the knowledge modeling of relational knowledge is crucial to enable reasoning systems to connect between two situations or domains. We investigate whether relational information is consistently modeled in Wikidata.
    \item \textit{Does Wikidata support extraction of analogy classification data?} Given its wide coverage, Wikidata may have the potential to generate large-scale analogy detection tasks automatically. We investigate how much manual effort is required to create such benchmarks for subclass-of relations in Wikidata, and we evaluate the performance of state-of-the-art NLP systems.
    \item \textit{Which desiderata and metrics can guide automatic generation of analogies using the Wikidata structure?} Considering that the Wikidata ontology is not uniform in terms of its granularity and expressivity, it is important to define desired properties for analogical reasoning and design automated metrics that can quantify this variation and enable the selection of potential analogical correspondences automatically.
\end{enumerate}
\section{Does Wikidata Express Relational Information Consistently?}
\label{sec:compositionality}

\subsection{Data and Setup} 
We sample 20 \textit{subclass-of (P279)} relations from Wikidata, whose subject label is a superstring of its object label. For example, we keep the subclass pair \textit{red wine - wine}, while we discard \textit{dog - pet}. We prioritize Qnodes with low identifiers as a simple proxy for well-known entities and concepts. Example pairs include \textit{stellar atmosphere - atmosphere} and \textit{computer keyboard - keyboard}.
We analyze whether the subclass-of relation is complemented by additional information that can help us categorize the nature of the inheritance relation, expressed either as other relations of the subject or qualifiers on the subclass-of relation.
Based on prior work on categorization of semantic relations for noun compounds~\cite{girju2005semantics}, we define an initial set of seven inheritance categories: PURPOSE, PROPERTY, LOCATION, OWNERSHIP, MATERIAL, INSTANCE, and TEMPORAL, and annotate each pair with one category. Besides obtaining relations for the original pair, we obtain siblings of the subject (other Qnodes that are direct children of the same object) and investigate their structures seeking for regularities. 

\begin{table}[t!]
\small
\caption{Ten exemplar compound noun pairs.}
\label{tab:compounds}
\begin{tabular}{ p{1.5cm} | p{1.7cm} | p{1.8cm} | p{1.6cm} | p{2cm} | p{3.9cm} }
\bf Subject & \bf Object & \bf Category & \bf Qualifiers & \bf Statements & \bf Siblings  \\ \hline
computer keyboard (Q250) & keyboard (Q1921606) & PURPOSE & follows: mobile phone & computer keyboard - part of - computer & typewriter keyboard, Braille keyboard, musical keyboard \\ \hline
natural science (Q7991) & science (Q336) & PROPERTY & -  & - & human science, information science, modern science, Ancient Egyptian science, ... \\ \hline
beach volleyball (Q4543) & volleyball (Q1734) & LOCATION & - & - & snow volleyball, women's volleyball, men's volleyball \\ \hline
fairy tale (Q699) & tale (Q17991521) & PROPERTY & - & - & old-fashioned tale, cumulative tale, urbain tale, German folk tale \\ \hline
Shia Islam (Q9585) & Islam (Q432) & INSTANCE & - & - & Sunni Islam, Islam in Denmark, Islamic eschaetology, Gospel in Islam \\ \hline
stellar atmosphere (Q6311) & atmosphere (Q8104) & LOCATION & of: star & - & extrasolar atmosphere, extraterrestrial atmosphere, Reducing atmosphere \\ \hline
electric charge (Q1111) & charge (Q73792) & PROPERTY & of: electromagnetic field & - & magnetic charge, color charge, weak hypercharge \\ \hline
red wine (Q1827) & wine (Q282) & PROPERTY & - & red wine - color - red & white wine, Mexican wine, Polish wine, straw wine, de-alcoholised wine, ...\\ \hline
day sky (Q4812) & sky (Q527) & PROPERTY & - & - & blue sky, morning sky, Velazquez sky, ...\\ \hline
animal rights (Q426) & right (Q2386606) & PROPERTY & of: nonhuman animal & - & hunting rights, women's rights, right to property, ... \\ \hline
\end{tabular}
\end{table}

\subsection{Findings}
We show ten out of the twenty pairs in Table~\ref{tab:compounds}.
Overall, we find that Wikidata describes relations sparsely, which does not help us identify the compound relation category. Specifically, out of 20 pairs, we found 4 cases where Wikidata provided a qualifier to further specify the relation. Among these four qualifiers, three were expressed with \textit{of} (e.g., \textit{stellar atmosphere - atmosphere} is further specified by the qualifier \textit{of - star}) and a single case used the \textit{follows} qualifier (\textit{computer keyboard - keyboard} is specified by \textit{follows - mobile phone}). In addition to being sparse, we find the qualifier information to correlate weakly with our semantic categories, as \textit{of} corresponds to both LOCATION (in \textit{stellar atmosphere - atmosphere - of: star}) and to PROPERTY (in \textit{electric charge - charge - of: electromagnetic field}).

Comparing the subject to its siblings also reveals a wide diversity in semantics that is not explicitly modeled. The siblings of natural science (Q7991) with a parent science (Q336) sometimes imply a property relation (e.g., human science, information science), and in others imply a spatial (science in Ivano-Frankivsk), temporal (modern science), or spatio-temporal (Ancient Egyptian science) specification. Some of the siblings with a PROPERTY relation use the \textit{studies} property in Wikidata to indicate the subject of the science (e.g., \textit{human science - studies - humans}). However, this property is again not consistently applied across the different siblings.
\section{Does Wikidata Support Extraction of Analogy Classification Data?} 
\label{sec:classification}

\subsection{Dataset construction}

Given that Wikidata does not support analogical reasoning directly through its relational modeling, we next investigate the possibility of creating a dataset of analogical pairs with Wikidata. Using the word pairs from the previous section as a seed set, we create analogies manually by searching for more P279 relations in Wikidata. We form 200 such quadruples, each consisting of two word pairs. The 200 quadruples are split evenly into analogical and non-analogical pairs.

Within the 100 positive cases (analogical pairs), we systematically sample three sets of cases: 1) 25 direct analogies (Pos-direct), which are pairs with a common object node, e.g., \texttt{computer science-science : food science-science}. 2) 50 parent-based analogies (Pos-parent), consisting of pairs of words where the objects have a common parent. For example, we sample the pair indoor \texttt{indoor golf-golf : road tennis-tennis}. 3) 25 distant analogies (Pos-distant), where the two objects have a more distant, yet semantically meaningful, common ancestor. An example is \texttt{Computer keyboard-keyboard : text display-display device}, where both keyboard and display device are products. For all three categories, we make sure that the analogical pairs are connected with the same relation, e.g., MATERIAL. To sample negative cases (Neg), we sampled random pairs and validated that their relation is not analogical. An example negative analogy is \texttt{Shia Islam-Islam : ancient music-music}. The sampling procedure of the positive and the negative analogical pairs included a manual postprocessing step, as Wikidata does not provide a direct way to filter the data. 

\begin{table}[t!]
\small
\centering
\caption{Distribution of the 100 positive analogies, with representative examples.}
\label{tab:dist}
\begin{tabular}{ c | r | c |  }
\bf Category & \bf Count & \bf Example  \\ \hline
PROPERTY & 53 & \texttt{fairy tale-tale : protestant cathedral-cathedral} \\ \hline
LOCATION & 19 & \texttt{sitting volleyball-volleyball : backyard cricket-cricket} \\ \hline
MATERIAL & 11 & \texttt{carbon fibers-fiber: iron wire-wire} \\ \hline
INSTANCE & 5 & \texttt{Shia Islam-Islam : Antiochian Greeks-Greek diaspora} \\ \hline
OWNERSHIP & 5 & \texttt{fetal liver-liver : elephant skin-skin} \\ \hline
PURPOSE & 4 & \texttt{computer keyboard-keyboard : scoreboard-display device} \\ \hline
TEMPORAL & 3 & \texttt{day sky- sky: winter garden-garden}  \\ \hline
\end{tabular}
\end{table}

\begin{table}[t!]
\small
\centering 
\caption{Examples of non-analogical pairs.}
\label{tab:neg_dist}
\begin{tabular}{ c | c | c   }
\bf Example & \bf Category 1 & \bf Category 2   \\ \hline
\texttt{beach basketball-basketball : Middle English-English} & LOCATION & TEMPORAL \\ \hline
\texttt{electronic music-music : Government of France-government} & PROPERTY & LOCATION \\ \hline
\texttt{carbon fibers-fiber : private bank-bank} & MATERIAL & PROPERTY \\ \hline
\texttt{mammal tooth-tooth - slang dictionary-dictionary} & OWNERSHIP & PROPERTY \\ \hline
\texttt{stellar atmosphere	atmosphere : marker pen-pen} & LOCATION & PURPOSE \\ \hline
\texttt{animal rights-right : cash-money} & PROPERTY & INSTANCE \\ \hline
\end{tabular}
\end{table}

The resulting 100 positive analogies are distributed as shown in Table~\ref{tab:dist}. This Table shows that some categories, most notably PROPERTY, dominate the dataset, suggesting that this category may need to be further refined in future work. In this Table, we also provide an example analogical pair for each of the seven categories an each of the three levels of analogy. We show negative examples in Table~\ref{tab:neg_dist}. Here, again the data is dominated by the PROPERTY relation.


\subsection{Baseline models}

Given that the dataset consists of proportional analogies on which language models have already been evaluated, we experiment with three models and to evaluate their performance: GloVe~\cite{pennington2014glove}, Word2Vec~\cite{mikolov2013efficient}, and BERT~\cite{devlin2018bert}. For GloVe and Word2Vec, we compute analogical similarity as $sim=cosine(emb(child1) - emb(parent1), emb(child2) - emb(parent2))$.
In the case of BERT, we compute similarity as $sim=cosine(emb(sentence1), emb(sentence2))$, where sentence1 and sentence2 are template-based sentences constructed following \textit{\{subject-label\} is \{subject-description\}}. This means that for BERT, we do not consider the parent nodes explicitly, they are only indirectly covered by the Wikidata descriptions of the subjects (children). 

For all models, we define a threshold $t$, such that the pairs are analogical if $sim>t$, and non-analogical otherwise. To rule out the impact of the threshold, we perform grid search and report the results with the optimal $t$.

\subsection{Results}

\begin{table}[t!]
\small
\centering
\caption{Baseline results on the analogical benchmark.}
\label{tab:results}
\begin{tabular}{ c | c | r }
\bf Model & \bf Supervised & \bf Accuracy  \\ \hline
GloVe & No & 0.72 \\
Word2Vec & No & 0.74 \\
BERT & No & 0.775 \\ 
BERT & Yes & \bf 0.815 \\ \hline
\end{tabular}
\end{table}

The best obtained results per model in are shown in Table~\ref{tab:results}. Among the unsupervised models, we observe that the performance of BERT is better than GloVe and Word2Vec. We also evaluate a supervised version of the BERT model, which is an SVM model trained with cross-validation. We observe that with this training, the BERT model is able to improve its performance to 81.5\%.

\begin{figure}[!t]
    \centering
    \includegraphics[width=0.98\linewidth]{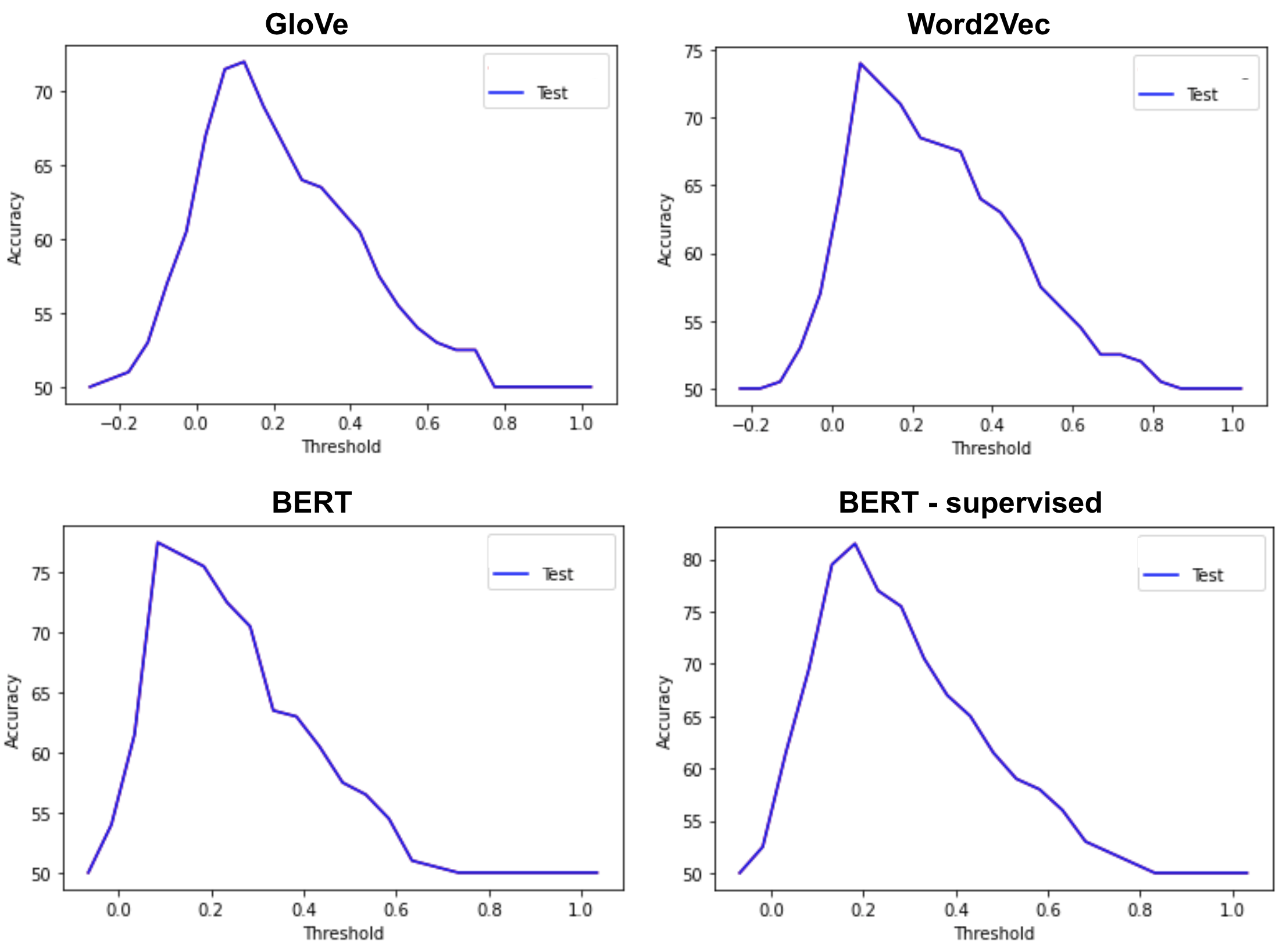}
    \caption{Optimal thresholds for each model.}
    \label{fig:threshold}
\end{figure}

\begin{figure}[!t]
    \centering
    \includegraphics[width=\linewidth]{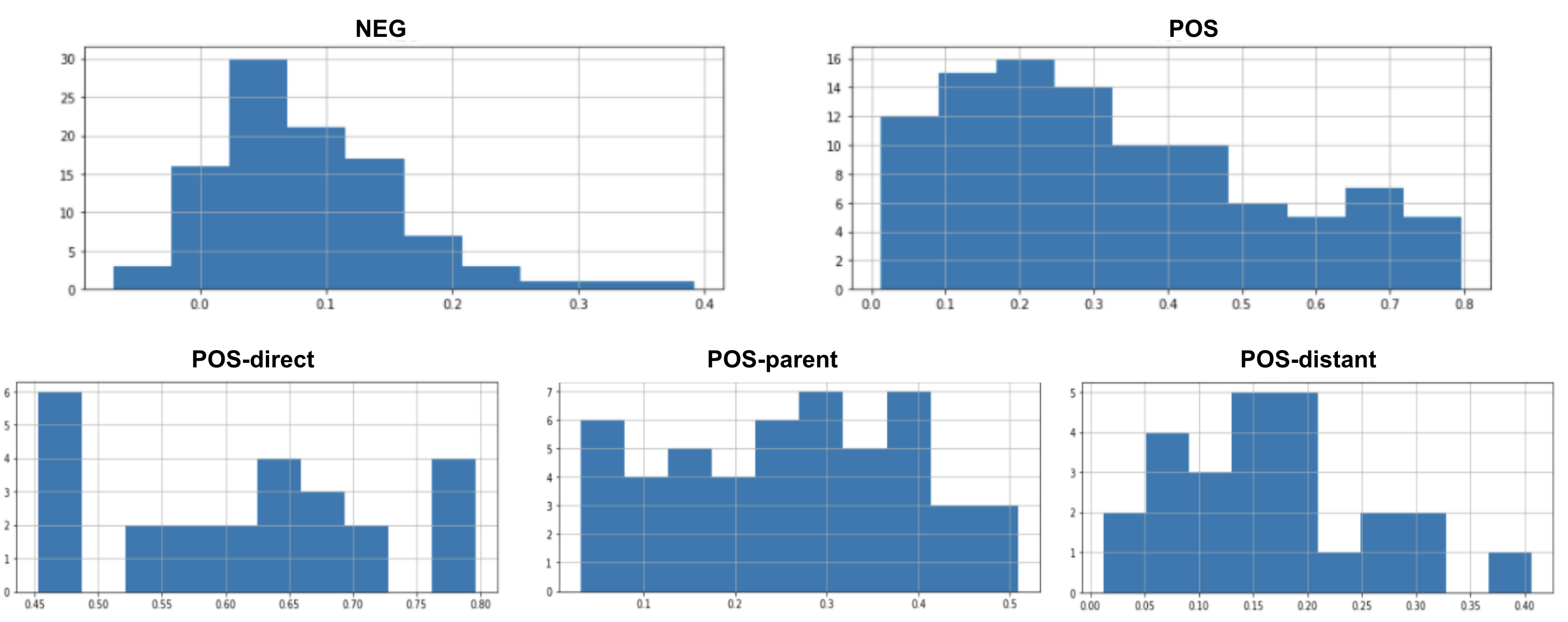}
    \caption{Distribution of cosine similarity values per analogy set.}
    \label{fig:counts}
\end{figure}

The impact of the different thresholds for each of the four models is shown in Figure~\ref{fig:threshold}. We observe that the optimal threshold for most models is around 0.1. Interestingly, when we distinguish between the similarities for the negative and the different kinds of positive cases (shown for BERT-supervised in Figure~\ref{fig:counts}), we observe that the similarities for the non-analogies are often between 0 and 0.1. The analogies for the positive class are more uniformly distributed, and peak between 0.2 and 0.3. Interestingly, the similarity values for both analogies and non-analogies are typically low, which signals that simple embedding calculations cannot be used to extract analogical links from language models, but also that the wide range of domains covered make this task even more challenging. 

Among the different positive classes, we observe that the direct analogies have highest similarities, and the similarity decreases with the ontological distance. These findings can be expected, and they indicate that language models contain some useful signals for identifying analogical pairs, yet they require further improvements to perform robustly. 

\begin{figure}[!t]
    \centering
    \includegraphics[width=0.55\linewidth]{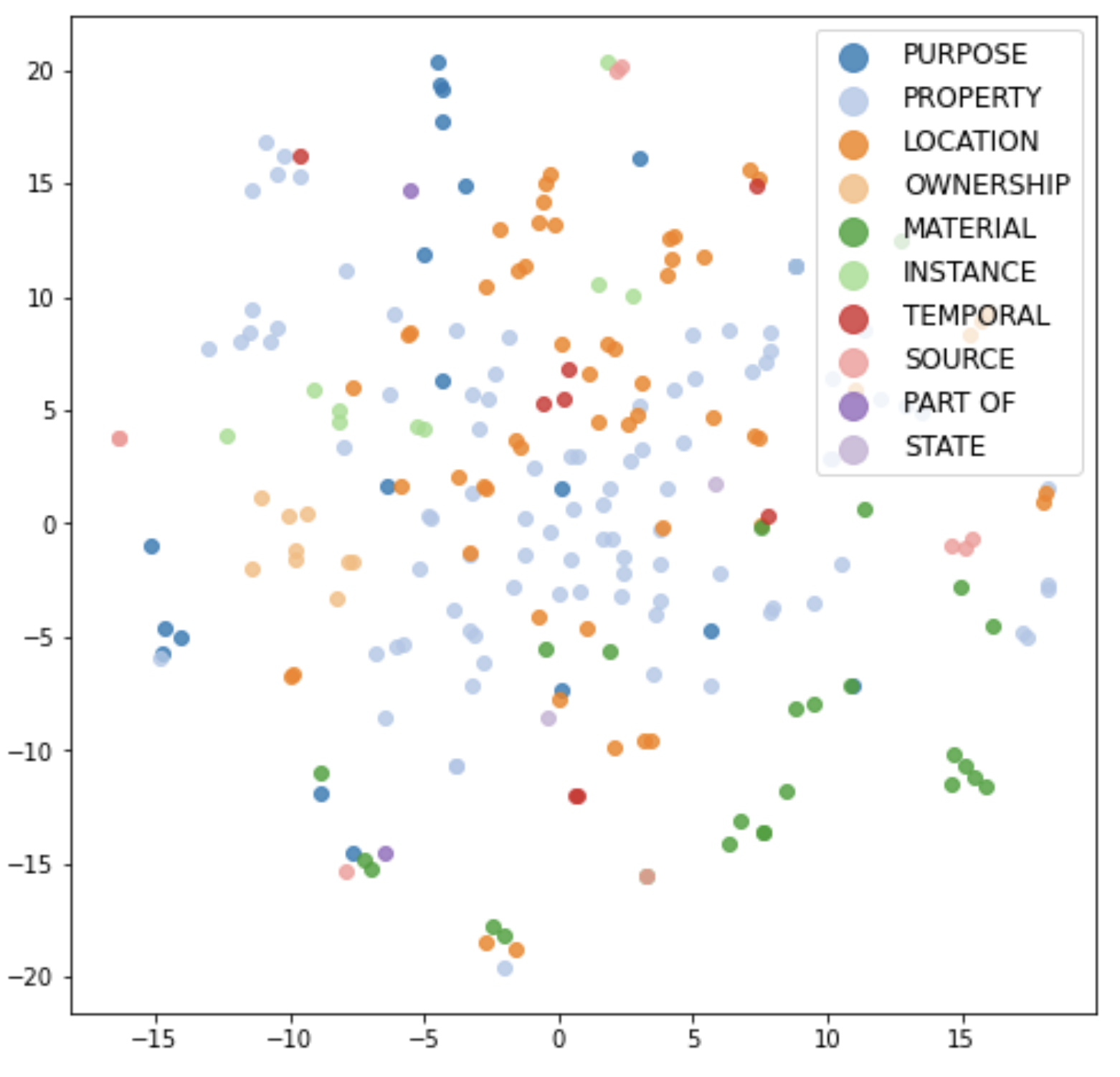}
    \caption{BERT-supervised embeddings for each pair, colored with its category.}
    \label{fig:relations}
\end{figure}

Considering these results, we investigate whether the models are able to distinguish between the different semantic categories of properties. Figure~\ref{fig:relations} shows a t-SNE plot of the BERT-supervised embeddings for each of the data points, colored by their category. The lack of same-colored visual clusters for BERT-supervised  shows that overall BERT is unable to clearly distinguish between the semantic categories. Some exceptions exist, for instance, the MATERIAL nodes are often found together in the bottom-right part of the plot. We obtained similar or worse results for the other models as well. This shows that even with supervision, language models are not able to directly understand analogical categories derived from Wikidata. These results are consistent with findings on existing datasets for analogical reasoning in the NLP domain~\cite{ushio-etal-2021-bert}.

\section{Towards Automated Analogy Generation Using Wikidata Structure}
\label{sec:metrics}
The experiments in the prior section were evaluated based on a small, manually curated set of analogical entities. To move beyond such manually curated resources, we explored some key desiderata for meaningful analogical relationships. Using the Wikidata ontology and a set of analytical metrics of ontological structure, we evaluated several techniques for filtering ontological classes to identify candidate analogical relationships. In this section, we describe the important properties of analogies, how such properties can be captured using metrics from the Wikidata ontology, and provide illustrative examples from five such approaches for analogy generation.

\subsection{What makes a good (Wikidata) analogy?}
Our exploration focused on classes in Wikidata, with the goal of finding meaningful child-parent relationships. We hypothesize that parent-child relationships provide a foundation for analogy generation since (a) child classes specialize from parent classes by some criteria and (b) sibling classes must differ from each other by some criteria. For example the class of houses is a subclass of buildings, and the differentiation between buildings and houses occurs based on purpose -- houses are buildings that are used for residential purposes, rather than commercial, governmental, or industrial purposes. Similarly, houses and office buildings are both types of buildings, but again differ based on purpose. To better understand the differences between parent and child classes and sibling classes, we consider entities that are instances of those classes. 

\paragraph{Generality:} Ontologies can contain very specific classes that are limited to a specialized domain, and used for technically precise concepts. For example, a butter mill is a type of building, but very few people have direct experience or understanding of butter mills. While such classes may still provide meaningful analogies, their specialization makes such analogies difficult to judge and less accessible. Our experiments exclude classes that may be too niche to provide a large set of meaningful analogies.
\paragraph{Selectivity:} Useful parent-child analogical class pairs must show a clear selective criteria for distinction between parent and child classes. If such class pairs are too similar, the conceptual differences between their constituent entities may be hard to characterize. For example, if most tools are devices, the distinction between tool and device diminishes. Non-selective entity pairs may result in confusing analogies, so useful analogical classes will show selective differences between classes.
\paragraph{Salience:} Similar to selectivity, salience is the ability to differentiate between two related classes. If two class pairs have largely overlapping constituent entities, the criteria to distinguish between classes may not be apparent. If many houses are also office buildings, the distinction between these two classes becomes difficult to justify. We choose salient pairs to ensure clear analogies. 
\paragraph{Diversity:} Interesting analogies may stem from parent classes that have several different subclasses that have significant instances. A structure with many meaningful subclasses can suggest a diverse set of concepts that can easily be used to create analogies. For example, by having many distinct building types with many different specific building instances, generating diverse analogies with buildings is more feasible. In contrast, if one building type dominates over others, or there are myriad, sparse building types analogy generation may become difficult.

\subsection{Metrics on the Wikidata ontology}
We pair the qualities of desirable ontological properties with specific metrics to realize these properties. By defining such metrics, we are able to analytically derive candidate classes and their associated instances for use in analogy generation. When paired with large scale validation data, these metrics can allow tailored analogy generation based on a specific type of analogy. We focus on two specific types of Wikidata relationships parent-child \texttt{subclass of} relationships (P279) and class-instance \texttt{is a} relationships (P31).

\paragraph{Generality $\rightarrow$ Instance Count:} To operationalize the generality of a Wikidata class, we introduce a metric of instance counts. 
Formally, $$IC = \setcard{C_i}$$ where IC is the instance count, $C_i$ is the set of instances of the class $C$ and
$\setcard{C_i}$ is the cardinality of set $C_i$.
A class with very few instances signals a specific or potentially undefined concept class. Filtering out classes with few instances can help avoid analogies that are difficult to understand.

\paragraph{Selectivity $\rightarrow$ Reduction Ratio:} We determine how selective a parent-child relationship is based on the ratio of child instances to parent instances, which we refer to as the reduction ratio. Formally,
$$ RR = 1-\frac{\setcard{C_i}}{\setcard{P_i}}$$
where $RR$ is the reduction ratio, $C_i$ is the set of instances of child class $C$ and $P_i$ is the set of instances of parent class $P$. 

If all parent instances are also child instances, the reduction ratio will be low. Low reduction ratios mean that a subclass is largely co-referent with the parent class and not very selective. In contrast, high reduction ratios mean that a subclass has very few instances of the parent class, and only represents a small portion of the concept space. Ideally, reduction ratios would be in between these extremes, identifying classes that still have a meaningful portion of the parent class without being too general or too specific.

\paragraph{Salience $\rightarrow$ Class Overlap:} To identify salience, we employ a ratio corresponding to class overlap as measured by the Jaccard set similarity. For each class, we measure the overlap with the largest sibling class. Formally, $$CO = \frac{\setcard{C_i \cap S_i}}{\setcard{C_i \cup S_i}}$$ $CO$ is the class overlap metric, $C_i$ are instances of class $C$ and $S_i$ are instances of sibling class $S$. The numerator designates the number of entities that are instances of both $C_i$ and $S_i$, while the denominator is the number of distinct entities that are instances of either $C$ or $S$. 

If the class overlap is high, the two classes share many instances and will be difficult to distinguish. If the class overlap is low, the two classes are largely distinct. Desirable analogical relationships between classes will require low class overlap to ensure that the classes are easily distinguished.

\paragraph{Diversity $\rightarrow$ Entropy:} We measure diversity of a parent class with respect to its subclasses by measuring the entropy. Formally, we first define the probability of a class as $P(C)=\frac{\setcard{C_i}}{\setcard{P_i}}$, where $P_i$ are instances of parent class $P$ and $C_i$ are instances of child class $C$. Next, we can formulate the entropy as $$H(P)=\prod_{C \subset P} -P(C)\cdot log(P(C))$$ $H(P)$ is the entropy of the parent class defined in terms of its subclasses $C$. 

If the entropy of a parent class is very low, the class hierarchy is likely to be dominated by a single, large subclass. Conversely, if the entropy is very high the hierarchy may contain a large number of small classes. A moderate value for entropies can achieve a balance between a dominating class and many small classes.

\subsection{Illustrative examples of Wikidata relationships}
We provide some illustrative examples found by filtering Wikidata with metrics above, using filters such as $IC > 1000, RR < .85, CO < 0.5, H(P) > 1$. One such example is the class of \textit{visual artwork} (Q4502142), which includes two sibling classes of \textit{painting} (Q3305213) and \textit{film} (Q11424). An example analogical pair at an instance level that can be generated from these two sibling classes is \texttt{The Bohemian} (Q1000128) :: \texttt{Am\'{e}lie} (Q484048), the former depicting and the latter having a narrative location of \texttt{Paris} (Q90). 

Another such example of sibling analogical relationships is for the parent class \textit{building} (Q3305213), which has subclasses \textit{residential building} (Q11755880) and \textit{religious building} (Q16970). An analogical pair at the instance level of these classes might be \texttt{Fallingwater} (Q463179) and \texttt{Unity Temple} (Q1680814), both of which were designed by architect \texttt{Frank Lloyd Wright} (Q5604).

 An example of a parent-child analogical relationship is for the parent class \textit{computer} (Q68) and the child class \textit{supercomputer} (Q121117). Instances that form an analogical pair are the \texttt{IBM 3790} (Q11223800) and Watson (Q12253), both of which were manufactured by \texttt{IBM} (Q37156).

These examples suggest that filtering the Wikidata ontology to identify classes containing high analogical potential is possible and can allow large-scale, automated analogies that capture interesting real-world relationships. In our ongoing work, we hope to develop fully automated analogies that can help explain the deeper structural relationships in the world using the immensity of knowledge in public knowledge graphs.
\section{Related Work}
\label{sec:relatedwork}

\textbf{Analogical reasoning} To perform analogical mapping and reasoning over structures, prior research has used publicly available portions from Cyc~\cite{lenat1995cyc}. For instance, the Companion architecture~\cite{Forbus2017AnalogyAQ} operates over a subset of Cyc integrated with several other resources, primarily stemming from the Natural Language Processing (NLP) domain. Most works on analogical reasoning in NLP have considered the task of establishing proportional analogies (a:b - c:d), where language models and word embeddings have been the dominant resource~\cite{mikolov2013efficient,ushio-etal-2021-bert}.
Nagarajah et al.~\cite{nagarajah2022understanding} studied the possibility of using language models or frames in FrameNet to perform various kinds of analogical reasoning over narratives.
In~\cite{wijesiriwardene2022towards}, the authors investigate distinguishing features that enable models to distinguish between weak and strong analogies in STEM.
While Wikidata has not been leveraged for analogical reasoning, recent work~\cite{das2022knowledge} has proposed a case-based reasoning method for Knowledge Base Question Answering that learns relational patterns in Wikidata between questions and their answers. Our study is complementary to prior work, revealing that Wikidata has much useful knowledge that can support analogies, yet it may not be ready yet for analogical reasoning at scale. Much of the knowledge that facilitates analogies is \textit{commonsense knowledge} (e.g., polio vaccine cures polio), whose coverage in Wikidata can be further improved~\cite{ilievski2020commonsense}. Interestingly, unlike other general-domain KGs like FreeBase,\footnote{\url{http://videolectures.net/iswc2017_taylor_applied_semantics/}, accessed October 2, 2022.} Wikidata's notability clause is flexible and supports the inclusion of commonsense knowledge.\footnote{\url{https://www.wikidata.org/wiki/Wikidata:Notability}}

\textbf{Quality and user experience}
Chen et al.~\cite{chen2019practical} devise a framework for evaluating the fitness of knowledge graphs for downstream applications. Piscopo and Simperl~\cite{piscopo2019we} survey dozens of papers and provide three categories of metrics: intrinsic (accuracy, trustworthiness, and consistency of entities), contextual (completeness and timeliness of resources), and representation (i.e., understanding, interoperability of entities). Enhanced metrics for timeliness of knowledge have been proposed by \cite{ferradji2021enhanced} and illustrated on a case study with Wikidata.
Noy et al.~\cite{noy2019industry} describe that large graph systems have three key determinants of quality and usefulness: coverage, correctness, and freshness, and investigate the approach towards achieving those at five major technical companies. The authors recommend consolidating descriptions of people, places, and other entities in Wikidata as a common core.
While Wikidata has been shown to be more reliable and expressive than other public knowledge graphs (e.g., DBpedia)~\cite{farber2018linked}, prior work has recognized challenges with its quality. Shenoy et al.~\cite{shenoy2021study} study over 300 Wikidata dumps and reveal that establishing identity, semantic typing, and satisfying semantic constraints are thorny issues that need further consideration. Recognizing that correctness, completeness, and freshness are difficult challenges, Wikidata has several tools to monitor, analyze, and surface issues with quality. These include the Objective Revision Evaluation Service (ORES) for vandalism detection,\footnote{\url{https://www.wikidata.org/wiki/Wikidata:ORES}} ReCoIn~\cite{10.1145/3184558.3191641} for completeness estimation, and the Primary Sources Tool (PST) for curation of the contribution process.\footnote{\url{https://www.wikidata.org/wiki/Wikidata:Primary_sources_tool\#References}} Our paper is orthogonal to prior efforts to organize, measure, or improve intrinsic aspects of Wikidata quality, as our goal is to investigate its fitness for analogical reasoning.
\section{Conclusions}
\label{sec:conclusion}

Recognizing that prior work on analogical reasoning has not considered the vast knowledge in Wikidata, this paper presented an initial study of the ability of Wikidata to support analogical reasoning. Our experiments with compositional subclass-of relations showed that relational knowledge in Wikidata is not consistently modeled, which can be expected given that Wikidata is created in a collaborative manner following the wisdom-of-the-crowd idea. Follow-up efforts to generate analogical classification data from Wikidata resulted in a dataset with 200 analogical quads (pair-of-pairs); yet, selecting these pairs required a substantial manual effort. To facilitate automatic analogy generation using the KG structure, we discussed what makes a good (Wikidata) analogy, suggesting four key desiderata: generality, selectivity, salience, and diversity. We paired these desirable ontological properties with specific metrics to realize them, and provided illustrative examples of Wikidata relationships sampled through these metrics. Future work will investigate how to apply the findings from this paper, and operationalize the desiderata and the metrics into a method that can sample analogical pairs. Future work should also investigate reframing and normalizing of the knowledge in Wikidata to allow for more direct support for analogical reasoning.
We release the data and the code supporting our analysis.
\footnote{\url{https://drive.google.com/file/d/1jJz4yAyBKjq4Mm47eMKD12w-DH5uugnN}}
\footnote{\url{https://github.com/usc-isi-i2/analogical-transfer-learning/blob/main/Analogical\%20Proj\%20Experiments.ipynb}}




\bibliographystyle{splncs04}
\bibliography{refs}
%


\end{document}